\documentclass[letterpaper,10pt,oneside,conference,final]{IEEEtran}

\usepackage{graphicx}
\usepackage{algorithm}
\usepackage{algpseudocode}
\usepackage{amssymb}
\usepackage{amsmath}
\usepackage{cite}
\usepackage{my_hyphenation}

\begin{document}

  \title{A Simple Reinforcement Learning Mechanism for Resource Allocation in LTE-A Networks with Markov Decision Process and Q-Learning}
  \author
  {
    \IEEEauthorblockN{Einar C. Santos\\}
    \IEEEauthorblockA{Federal University of Goias\\
                  Av. Dr. Lamartine Pinto de Avelar, 1120\\
                  Catalao - GO - Brazil\\
                  einar@ufg.br}
  }
  \maketitle

  \begin{abstract}
	Resource allocation is still a difficult issue to deal with in wireless networks. The unstable channel condition and traffic demand for Quality of Service (QoS) raise some barriers that interfere with the process. It is significant that an optimal policy takes into account some resources available to each traffic class while considering the spectral efficiency and other related channel issues. Reinforcement learning is a dynamic and effective method to support the accomplishment of resource allocation properly maintaining QoS levels for applications. The technique can track the system state as feedback to enhance the performance of a given task. Herein, it is proposed a simple reinforcement learning mechanism introduced in LTE-A networks and aimed to choose and limit the number of resources allocated for each traffic class, regarding the QoS Class Identifier (QCI), at each Transmission Time Interval (TTI) along the scheduling procedure. The proposed mechanism implements a Markov Decision Process (MDP) solved by the Q-Learning algorithm to find an optimal action-state decision policy. The results obtained from simulation exhibit good performance, especially for the real-time Video application.
  \end{abstract}


\section{Introduction}

	Wireless networks are remarkably known due to its unpredictable physical conditions. Its channel suffers intense variation caused by numerous aspects: signal pathloss; fading; \emph{et cetera}. Additionally, is critical nowadays the offering of Quality of Service (QoS) support for applications since the resource demand is actually becoming far more stringent.

	The machine learning is emerging as an attractive choice among the variety of techniques suited to optimize resource allocation for recent wireless networks. Reinforcement learning is particularly unique to help the achievement of an optimal performance by the system orienting it from a resultant output after a performed action.

	In this context, it is proposed a reinforcement learning mechanism for Long Term Evolution Advanced (LTE-A) networks designed to determine and restrict the number of Resource Blocks (RBs) available for each traffic class. The proposal models the problem of choosing the number of RBs as a Markov Decision Process (MDP) and is solved running the Q-Learning algorithm.

	The mechanism is attractive because of its simplicity. The Q-Learning algorithm is simple, straightforward and efficient to solve finite state MDPs. Concerning complexity, it is computationally cheap and easy to implement \cite{sutton2011reinforcement}.

	In fact, the MDP is a popular tool for modeling agent-environment interaction \cite{wang2016survey}. Several works in the literature implement MDP in wireless networks for various applications, raising the interest in the concept and its comprehensiveness. In a MDP every decision corresponds to an action taken towards a represented state of the process, helping the system to evaluate its condition.

	The proposed mechanism directs its effort in analyzing the system state subject to the QoS parameters of applications as throughput, delay, and packet loss rate, instead of evaluating physical information of the wireless system.

	The mechanism is evaluated at the system level through simulation and compared it with some scheduling algorithms utilized in LTE-A networks.

	This paper is arranged as follows: in Section \ref{related_work} there is some discuss some of the related works and their resemblances with the proposal; the fundamentals of MDPs are presented in Section \ref{markov_decision_process}; the description of Q-Learning algorithm is shown in Section \ref{q-learning}; in Section \ref{mechanism_description} the proposed mechanism is exposed; simulation parameters are shown in Section \ref{simulation_parameters} while results are presented in Section \ref{results} with some discussion; finally, in Section \ref{conclusion} it brings some conclusion about the whole work.

\section{Related Work}
\label{related_work}

	In \cite{yu2008new, yu2004new} the authors propose a reinforcement learning method aimed to improve QoS provisioning for adaptive multimedia applications in cellular wireless networks defining policies for Call Admission Control (CAC) and Bandwidth Adaptation (BA). They adopted Semi-Markov Decision Process (SMDP) -- which treats continuous-time problems as discrete-time \cite{sutton2011reinforcement} -- to model and solve the problem. It is important to take into account the CAC procedure in order to control the system resources, but this also can be accomplished at the resource allocation level.

	The authors in \cite{shahid2015docitive} devised a distributive reinforcement learning mechanism for joint resource allocation and power control on femtocell networks. Each femtocell seeks to maximize its capacity while maintaining QoS. The Q-Learning algorithm is adopted, and the information about each independent learning procedure is shared among the femtocells to speed up the overall learning process. However, they not consider traffic differentiation and some action to avoid the services may harm each other.

	The Scheduling-Admission Control (SAC) for a generic wireless system is appropriately investigated in \cite{phan2013optimal}. Authors also propose two online learning algorithms in order to optimize the SAC procedure with low complexity and convergence faster than the Q-Learning algorithm. They approached the problem with a model-based solution, however, ignoring QoS issues. In contrast, it is important to implement the model-free approach because it can cover any technology in the field, independently.

\section{Markov Decision Process}
\label{markov_decision_process}

	The MDP is a control process model with stochastic, memoryless and discrete time properties. It is formally described as a tuple \(\left<S, A, P, R\right>\) where \(S\) is the state set, \(A\) is the action set, \(P\) is a set comprising the transition probabilities among states, and \(R\) is a reward set containing a \(r\) value for each action \(a\) taken \cite{white1989markov, sigaud2013markov, pellegrini2007processos}. Each \(p\) value, with \(p \in P\), measures the probability of an action \(a \in A\) be performed at a decision epoch. An action \(a\) changes the process state from a \(s\) to a new \(s'\) value and represents a decision-making.

\begin{figure}[!htb]
		\centering
		\includegraphics[width=0.3\textwidth]{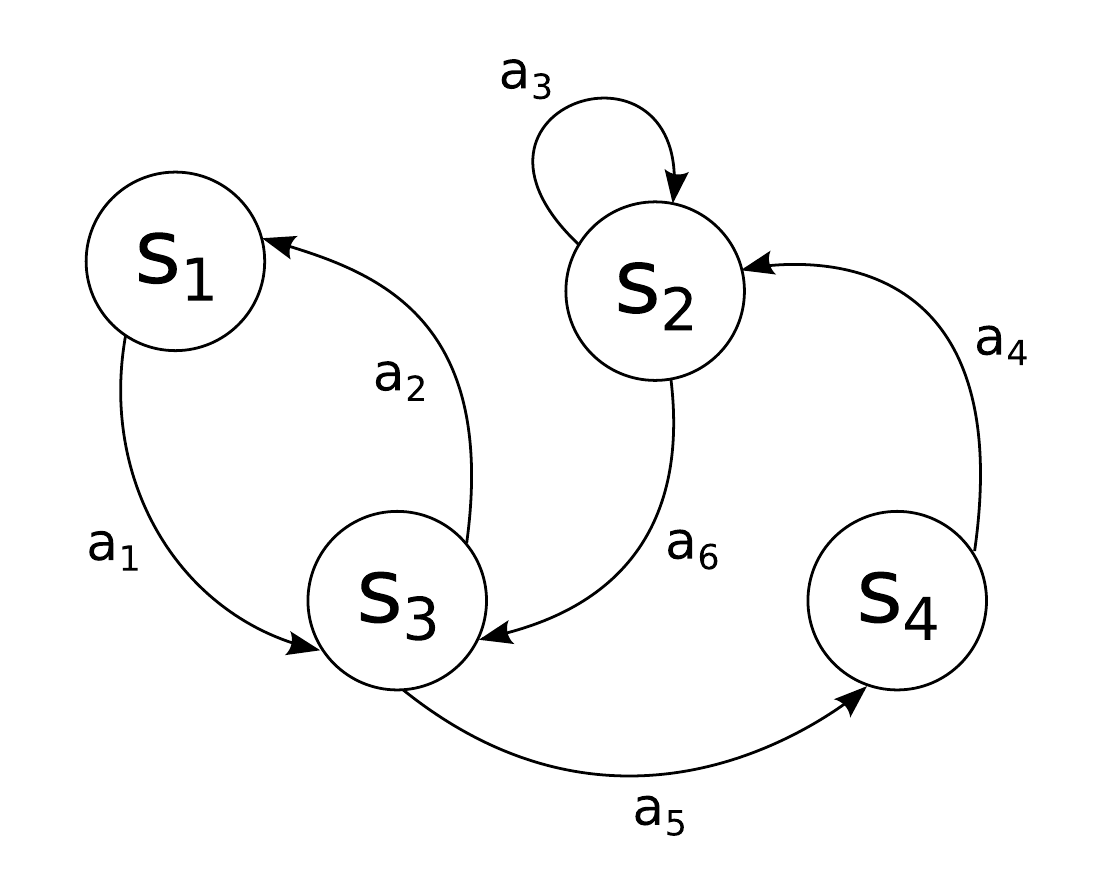}
		\caption{Markov Decision Process with finite states}
		\label{mdp}
	\end{figure}

	Figure \ref{mdp} depicts a MDP according to the given definition. For simplicity, transition probabilities \(P\) and the reward values \(R\) are not displayed.
	
	The decision epoch is a discrete time unit adopted for decision-making. If the number of decision epochs is finite, the MDP formulation is referred as finite-horizon. The horizon also can be infinite or undefined, when a MDP stops if a final state is reached.

	A decision taken at an epoch \(k\) is a function \(d_{k}\) that maps the rule \(d_{k}(s): S \mapsto A\) which also corresponds to an action ($d_{k}(s) = a_{k}$) \cite{pellegrini2007processos}. A policy \(\pi\) is defined as the collection of decision rules, given as: \(\pi = \{d_{0}, d_{1}, \dots, d_{Z-1}\}\). The variable \(Z\) is the total number of decision epochs.

	An optimal policy is the one that maximizes the measure of long-run expected rewards. It can be obtained from the optimal value of total reward function \(u^{*}_{k}(s)\), given as follows \cite{alagoz2010markov}:

	\begin{equation}
		\begin{aligned}
			u^{*}_{k}(s_{k}) &= \max_{d_{k}(s) \in A}{ \Big\{r_{k}\big(s_{k},d_{k}(s)\big)} \\
					 &{+ \gamma\sum_{s' \in S}^{}{p\big(s_{k}, d_{k}(s), s'\big)u_{k+1}^{*}(s')} \Big\}}
		\end{aligned}
	\end{equation}

	The \(\gamma\) value is the discounting factor used to weight immediate rewards.

	In order to store values for every state-action pair the MDP should consider the \(Q(s,a)\) function, usually called action-value function or simply \(Q\)-function \cite{sutton2011reinforcement}:

	\begin{equation}
		Q(s,a) = r(s,a) + \gamma \max_{a'}{Q(s',a')}
	\end{equation}

\section{Q-Learning Algorithm}
\label{q-learning}
	
	Q-Learning \cite{watkins1992q, watkins1989learning} is a straightforward and model-free reinforcement learning algorithm adopted to define the values of transition probabilities and to find an optimal policy for a MDP. It also converges to an optimal policy given a finite action-state MDP \cite{jaakkola1994convergence, melo2001convergence}.

	The Q-Learning algorithm iteratively updates the \(Q\)-table for each $(s,a)$ pair visited. The formula for updating the values at each step \(t\), with the learning rate $\alpha$, is:

	\begin{equation}
		\begin{aligned}
		\label{calculo_qlearning}
			Q(s_{t},a_{t})	&= Q(s_{t},a_{t}) \\
					&+ \alpha \left[r_{t+1} + \gamma \max_{a}{Q(s_{t+1},a)} - Q(s_{t},a_{t})\right]
		\end{aligned}
	\end{equation}

	The algorithm is given as follows \cite{sigaud2013markov}:

	\begin{algorithm}
		\caption{Q-Learning Algorithm}\label{q_learning_algorithm}
		\begin{algorithmic}[1]
			\State Initialize $Q_{0}$ and $\alpha$
			\For{$t = 0$ \textbf{to} $T-1$} 
				\State Select state $s_{t}$
				\State Select action $a_{t}$
				\State Send $a_{t}$ and $s_{t}$ information to the environment
				\State Get reward value $r_{t+1}$
				\State Calculate $Q(s_{t},a_{t})$ according to (\ref{calculo_qlearning})
			\EndFor
		\end{algorithmic}
	\end{algorithm}

	In a deterministic model, to ensure that all state-action pair is going to be visited, the algorithm must randomly select the \((s_{t},a_{t})\) pair and run the \emph{for} loop during a sufficient total number of steps \(T\) previously chosen. However, considering the case of a stochastic model (with unpredictable reward values), a nice option is to implement \emph{online} the \(\epsilon\)-greedy method, which selects the \((s,a)\) pair randomly with $\epsilon$ probability, balancing the system and finding a way to circumvent the \emph{exploitation versus exploration} dilemma \cite{sutton2011reinforcement}. The second option is also suitable when the system will run during an infinite (or unknown) number of steps.

\section{Mechanism Description}
\label{mechanism_description}

	The proposed mechanism is just referred as MDP with Q-Learning (MDP-QL).
	
	For QoS guarantee the system should look at the traffic QoS Class Identifier (QCI) priority values \cite{3gppTS23203} and also to classification from these values. The number of resources is selected from the current state of MDP at each Transmission Time Interval (TTI). Meaning that every MDP state denotes a portion of available RBs and one TTI is equivalent to one decision epoch.

	If the limit of available resources is reached for a traffic class, the system must advance to the next class observing QCI priority values until all resources or traffic have been exhausted in allocation procedure. RB metrics for every User Equipment (UE) are calculated as usual but restricting the usable load regularly for each class according to the selected proportion.

	Firstly, the state-action table \(Q(s,a)\) should be created and initialized for the MDP so that the system can perform the selection from state-action values. At the early steps the chosen values will not serve nicely for the needed quantity, but as long as the algorithm runs the state-action table, it will converge to more appropriate values.

	\subsection{Model Definition}
	\label{model_definition}

		The MDP adopted in this proposal has a finite number of states \(\Sigma\) and is infinite-horizon once the mechanism should operate indefinitely. However, for simulation purposes the horizon can be assumed as finite, so the number of decision epochs is constrained by the simulation time.

		Of course, the \(\Sigma\) value must be carefully chosen. A large number of states consequently makes the running impracticable. This is also referred as the curse of dimensionality \cite{alagoz2010markov}.

		Each state \(s_{k,i}\) contains the value that delimitates the maximum number of RBs for a treated traffic class. The total number of RBs \(N\) in the system, the state index \(i\) and the total number of states \(\Sigma\) settled should be considered:
	
		\begin{equation}
			s_{k,i}=\left\lceil \frac {N \cdot i}{\Sigma} \right\rceil \quad i =\{1,2,\dots,\Sigma\}, \quad N > \Sigma.
		\end{equation}

		In order to obtain reward at some decision epoch, the mechanism has to monitor some indicators. In this proposal the goal is to improve the system as follows:

		\begin{equation}
			r_{k} = \log{\left(\frac{\bar{R_{k}}}{\bar{\delta_{k}} \bar{\rho_{k}}}\right)}
		\end{equation}

		The reward value \(r_{k}\) takes into account the average throughput \(\bar{R_{k}}\), the average delay \(\bar{\delta_{k}}\) and the average packet loss rate \(\bar{\rho_{k}}\) of all running applications in the system at the decision epoch \(k\). Thus, it is established as overall goal the system throughput maximization while reducing its delay and packet loss rate. The $\log\left(\right)$ function is used to compensate the scale.
	
\begin{figure}[!htb]
			\centering
			\includegraphics[width=0.33\textwidth]{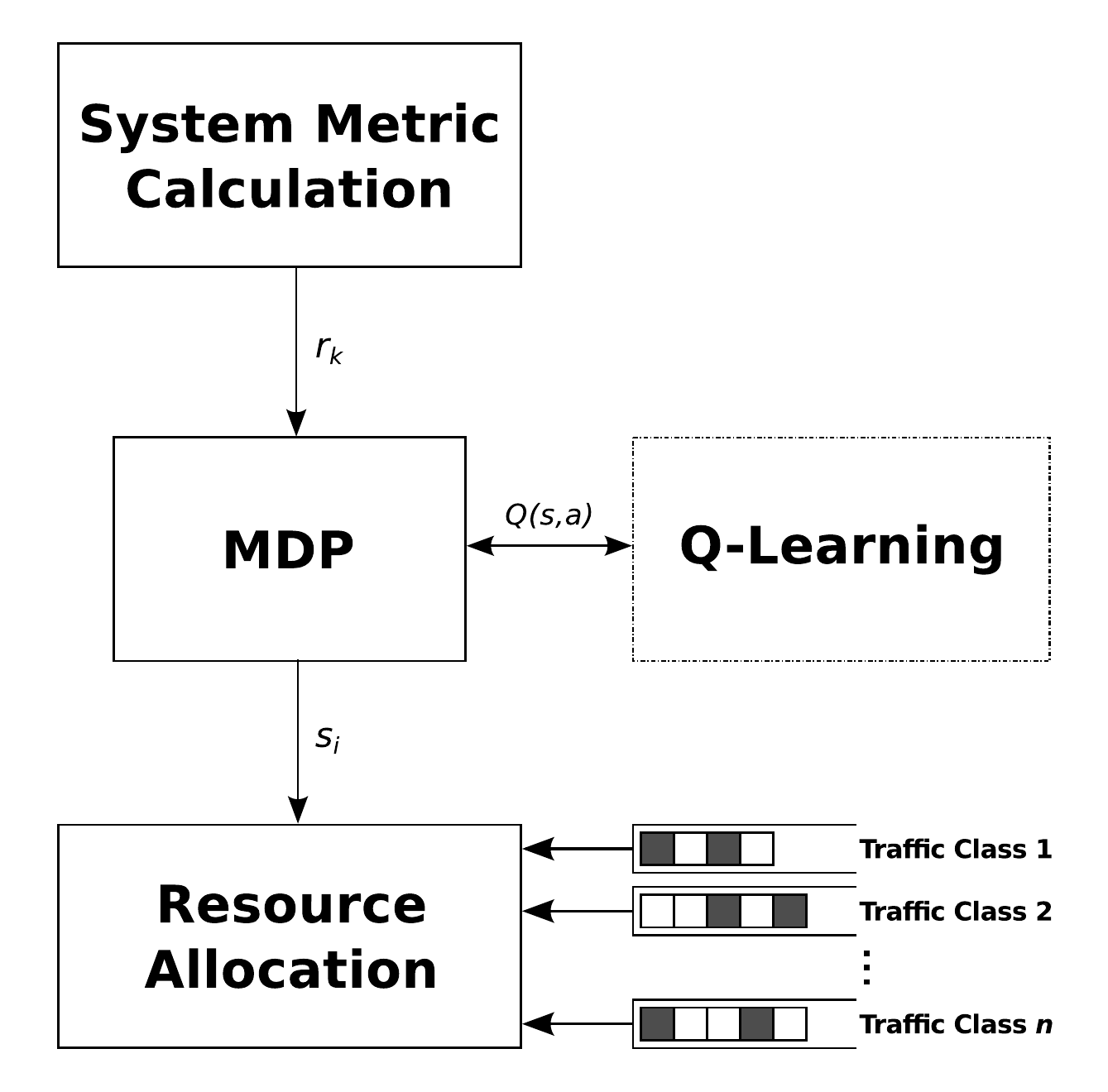}
			\caption{Scheme of the proposed mechanism}
			\label{mechanism_scheme}
		\end{figure}

		The mechanism scheme is depicted in Figure \ref{mechanism_scheme}.

\section{Simulation Parameters}
\label{simulation_parameters}

	This Section is dedicated to present the values of selected parameters contemplated in simulation definition to perform mechanism evaluation at the system level. In this particular case solely the downlink channel was analyzed. However, the mechanism can be studied from the uplink channel perspective as well. The parameters values are presented in Table \ref{parametros_de_simulacao}.
	
	Every UE is randomly positioned in the cell before simulation starting and has one running instance of each application: real-time Video, VoIP, and Web.

	\begin{table}[!htb]\footnotesize
	\centering
	\caption{Simulation parameters}
		\begin{tabular}{ c | c }
			\hline
			\textbf{Parameter}			& \textbf{Value / Description}			\\ \hline\hline
			Modulation				& OFDMA						\\ \hline
			Frequency				& 2 GHz						\\ \hline
			Bandwidth				& 20 MHz					\\ \hline
			Cell Radius				& 1 Km Single Cell				\\ \hline
			Frame Duration				& 1 ms						\\ \hline
			Duplexing Mode				& FDD						\\ \hline
			Power - eNB				& 46 dBm					\\ \hline
			Power - UE				& 23 dBm					\\ \hline
			Pathloss Model				& Urban (\(P_{L} = 128.1 + 37.6 \log{d}\))	\\ \hline
			Simulation Time				& 60 s						\\ \hline
			Traffic Time				& 54 s						\\ \hline
			Number of UE				& 10 -- 120					\\ \hline
			UE Speed				& 30 Km/h					\\ \hline
			Application 				& Video, VoIP and Web				\\ \hline
			Video					& H.264 440 Kbps				\\ \hline
			VoIP					& G.711 64 Kbps					\\ \hline
			Web					& Pareto					\\ \hline
			Video Delay Limit			& 150 ms					\\ \hline
			VoIP Delay Limit			& 100 ms					\\ \hline
			\(\Sigma\) Value			& 20						\\ \hline
			\(\alpha\) Value			& 0.2						\\ \hline
			\(\gamma\) Value			& 0.75						\\ \hline
			\(\epsilon\) Value			& 0.1						\\
			\hline
		\end{tabular}
	\label{parametros_de_simulacao}
\end{table}

	It was chosen the following scheduling algorithms for comparison: Proportional Fair (PF); Round Robin (RR) and Frame Level Scheduling (FLS) \cite{piro2010two}.

\section{Results}
\label{results}

	The following results present the behavior of evaluated algorithms including the proposed mechanism. It was analyzed throughput, delay, jitter, fairness index and packet loss.

\begin{figure}[!htb]
		\centering
		\includegraphics[width=0.4\textwidth]{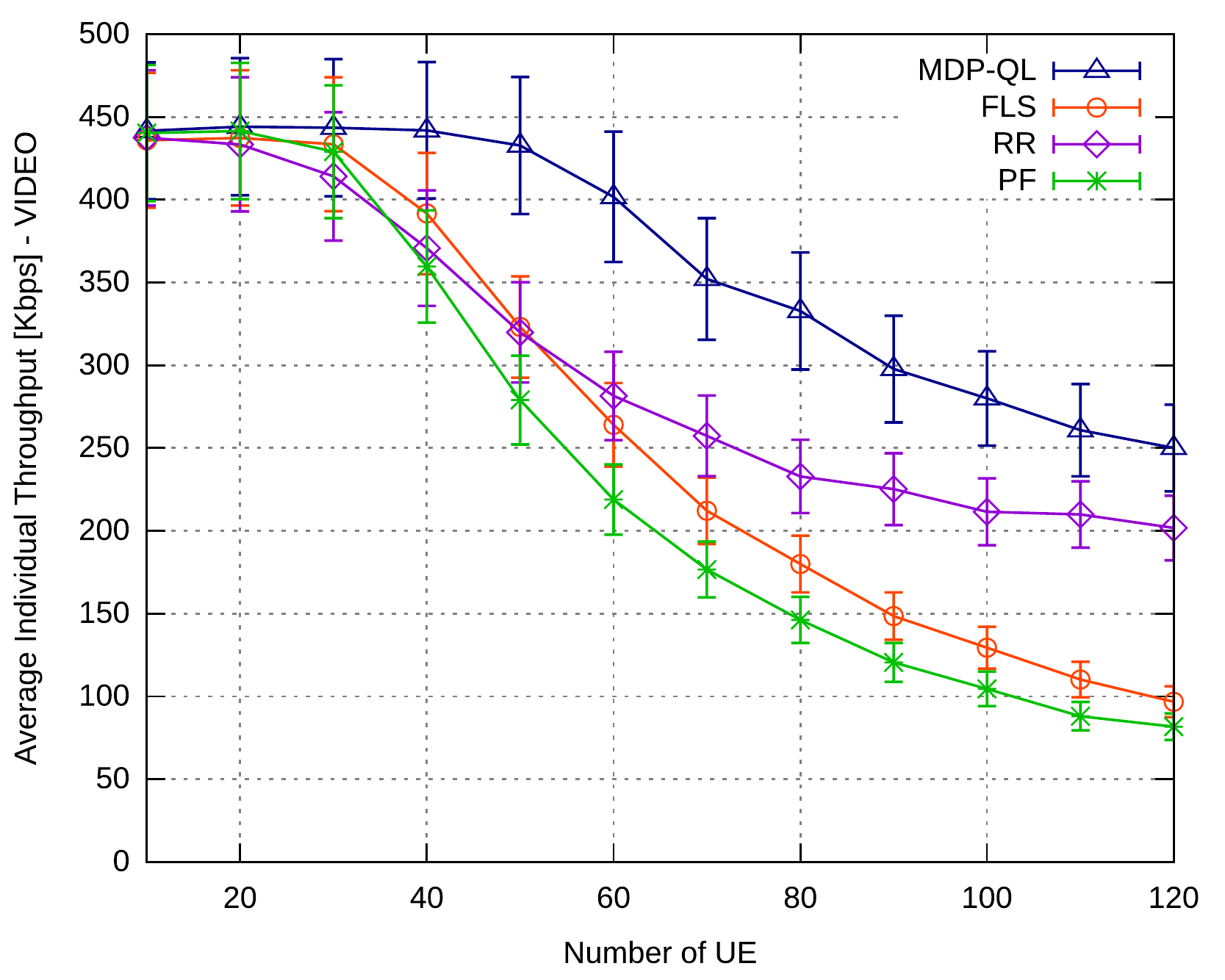}
		\caption{Average throughput for Video application}
		\label{throughput_video}
	\end{figure}

	Figure \ref{throughput_video} presents the throughput performance obtained for the Video application.

\begin{figure}[!htb]
		\centering
		\includegraphics[width=0.4\textwidth]{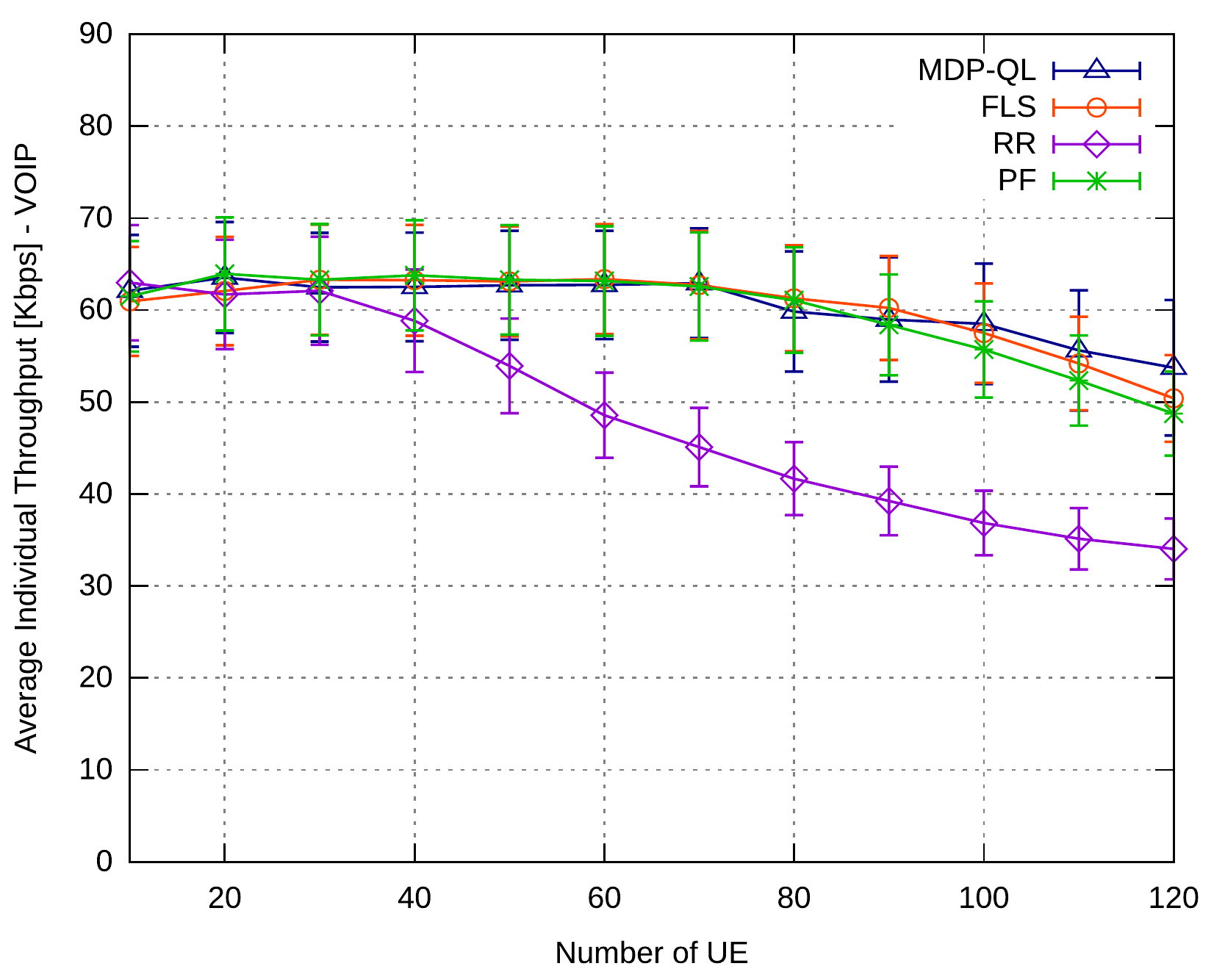}
		\caption{Average throughput for VoIP application}
		\label{throughput_voip}
	\end{figure}

	The average throughput for Video application running MDP-QL maintains the needed rate for application up to 50 UEs. The MDP-QL also keeps the average throughput higher than the other algorithms when the system has more than 40 UEs. To ensure that this performance is not occurring to the detriment of the VoIP traffic the Figure \ref{throughput_voip} shows the average throughput for VoIP application is still preserved with MDP-QL as well as in the other algorithms up to 70 UEs.

\begin{figure}[!htb]
		\centering
		\includegraphics[width=0.4\textwidth]{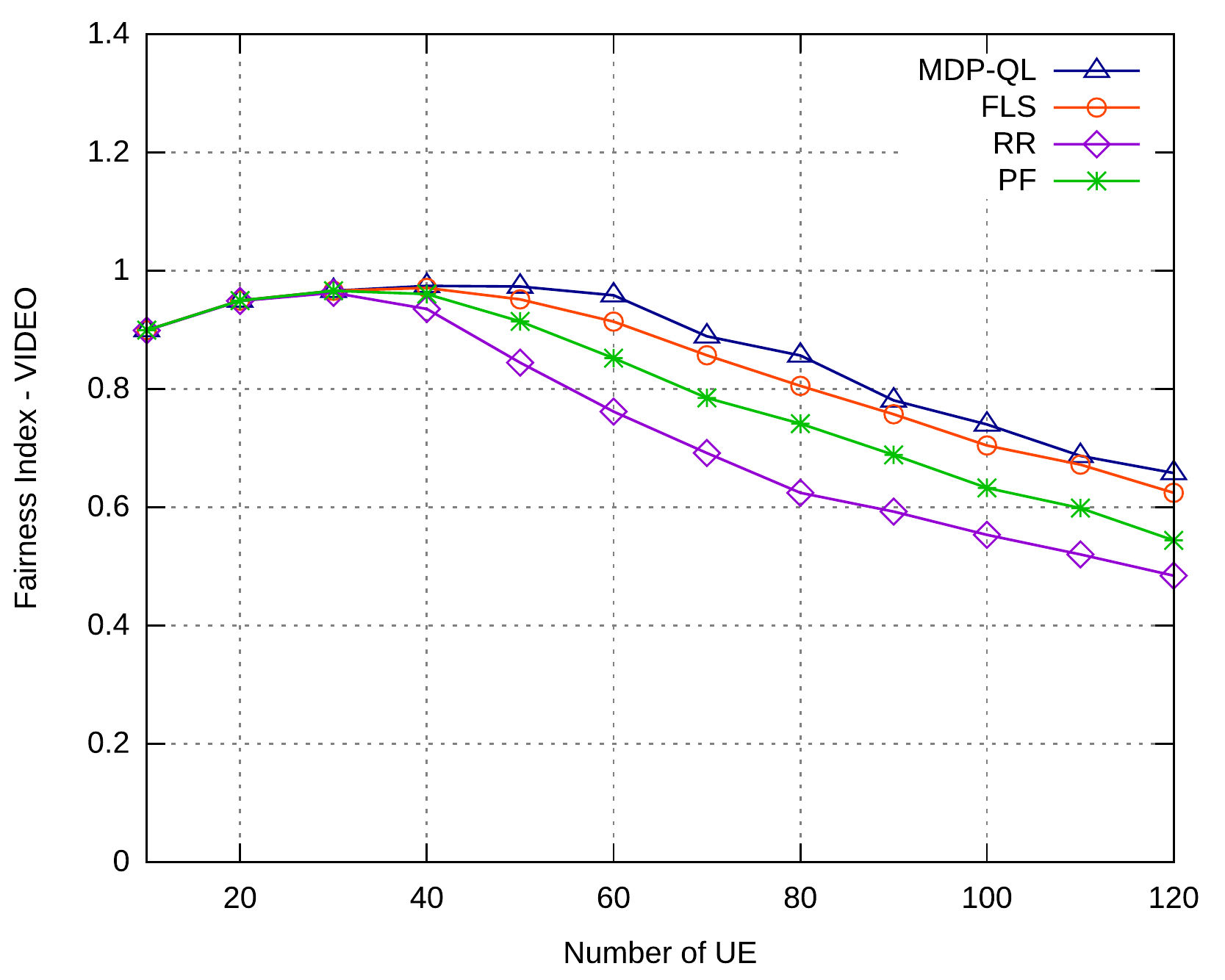}
		\caption{Fairness index for Video application}
		\label{fairness_index}
	\end{figure}

\begin{figure}[!htb]
		\centering
		\includegraphics[width=0.4\textwidth]{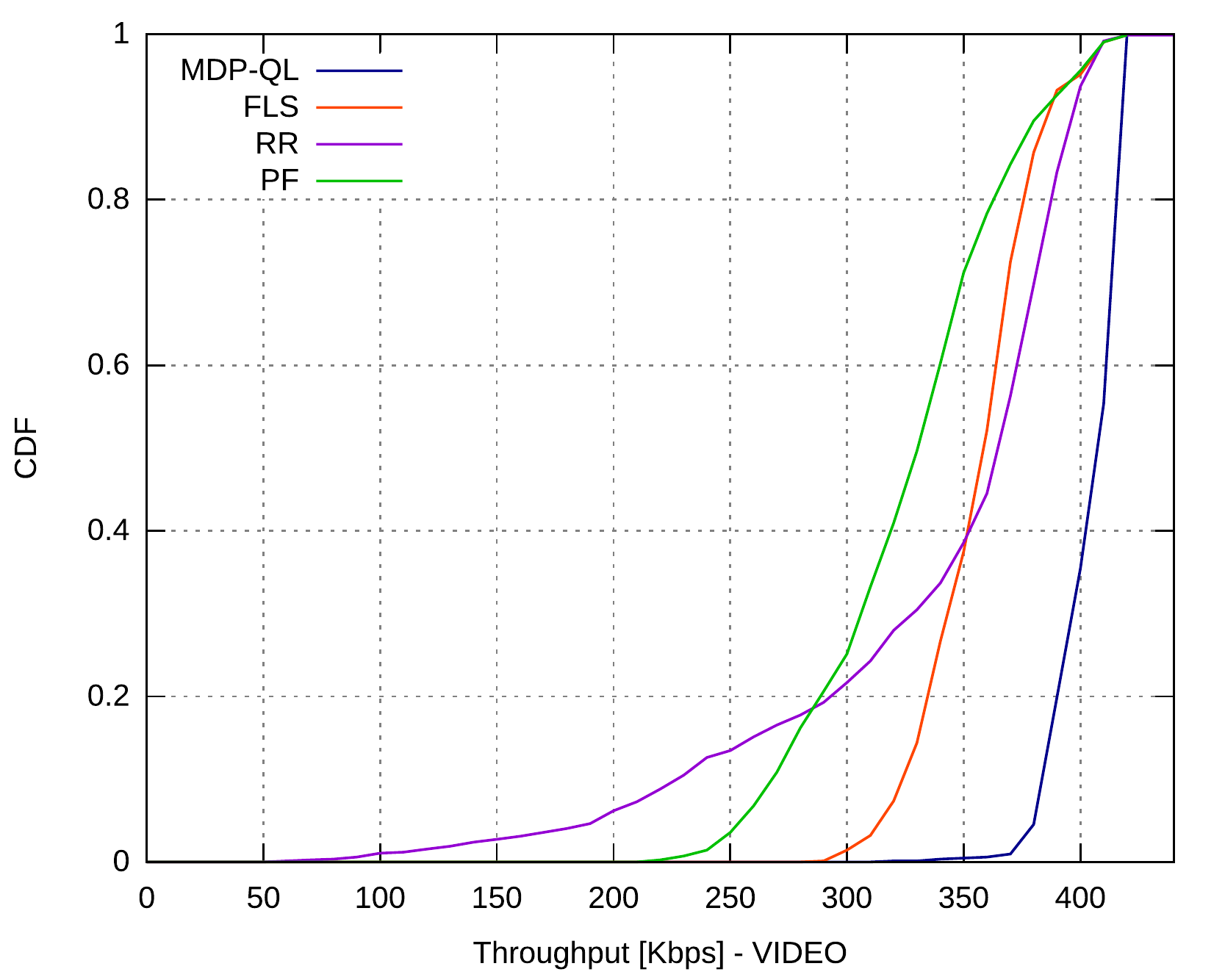}
		\caption{CDF for Video application with 40 UEs}
		\label{cdf_throughput_40_ues}
	\end{figure}

	When analyzing the throughput sharing among UEs is possible to see in Figure \ref{fairness_index} that the fairness index for MDP-QL is above 0.9 up to 60 UEs in the system and slightly above the other algorithms from 60 UEs, too. It means that MDP-QL, in addition to maintaining a good throughput, still balances it in terms of sharing. Such statement is corroborated by results presented in Figure \ref{cdf_throughput_40_ues}, showing the curves of Cumulative Distribution Function (CDF) for average throughput with 40 UEs in the system. The MDP-QL allocates just over 60\% of UEs with throughput higher than 400 Kbps.

\begin{figure}[!htb]
		\centering
		\includegraphics[width=0.4\textwidth]{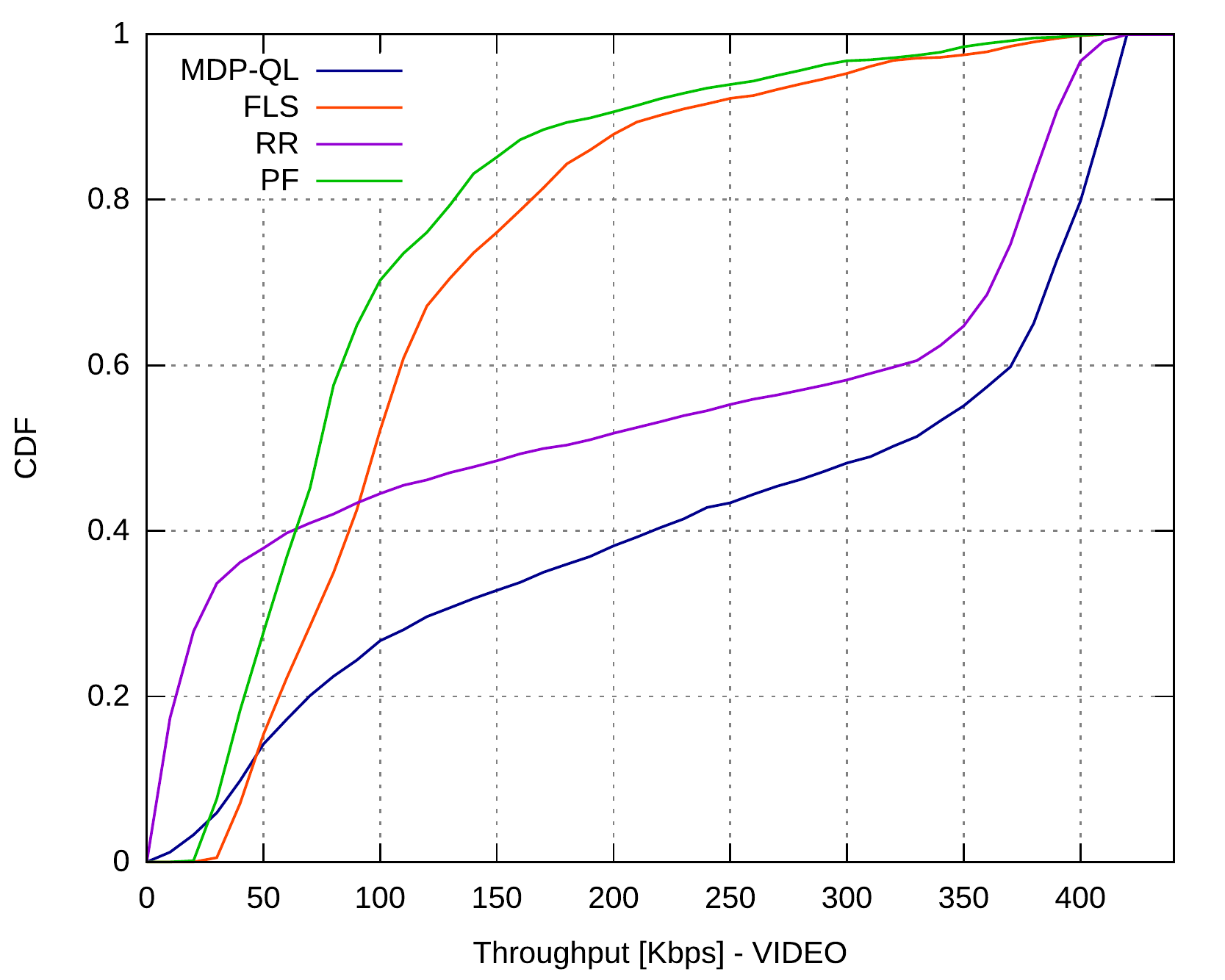}
		\caption{CDF for Video application with 100 UEs}
		\label{cdf_throughput_100_ues}
	\end{figure}

	Figure \ref{cdf_throughput_100_ues} exhibit CDF for average throughput with 100 UEs present. The graphic in Figure \ref{cdf_throughput_100_ues} is useful to demonstrate the behavior of MDP-QL and its evolution from 40 UEs, which still conserves good distributed throughput in such condition.

\begin{figure}[!htb]
		\centering
		\includegraphics[width=0.4\textwidth]{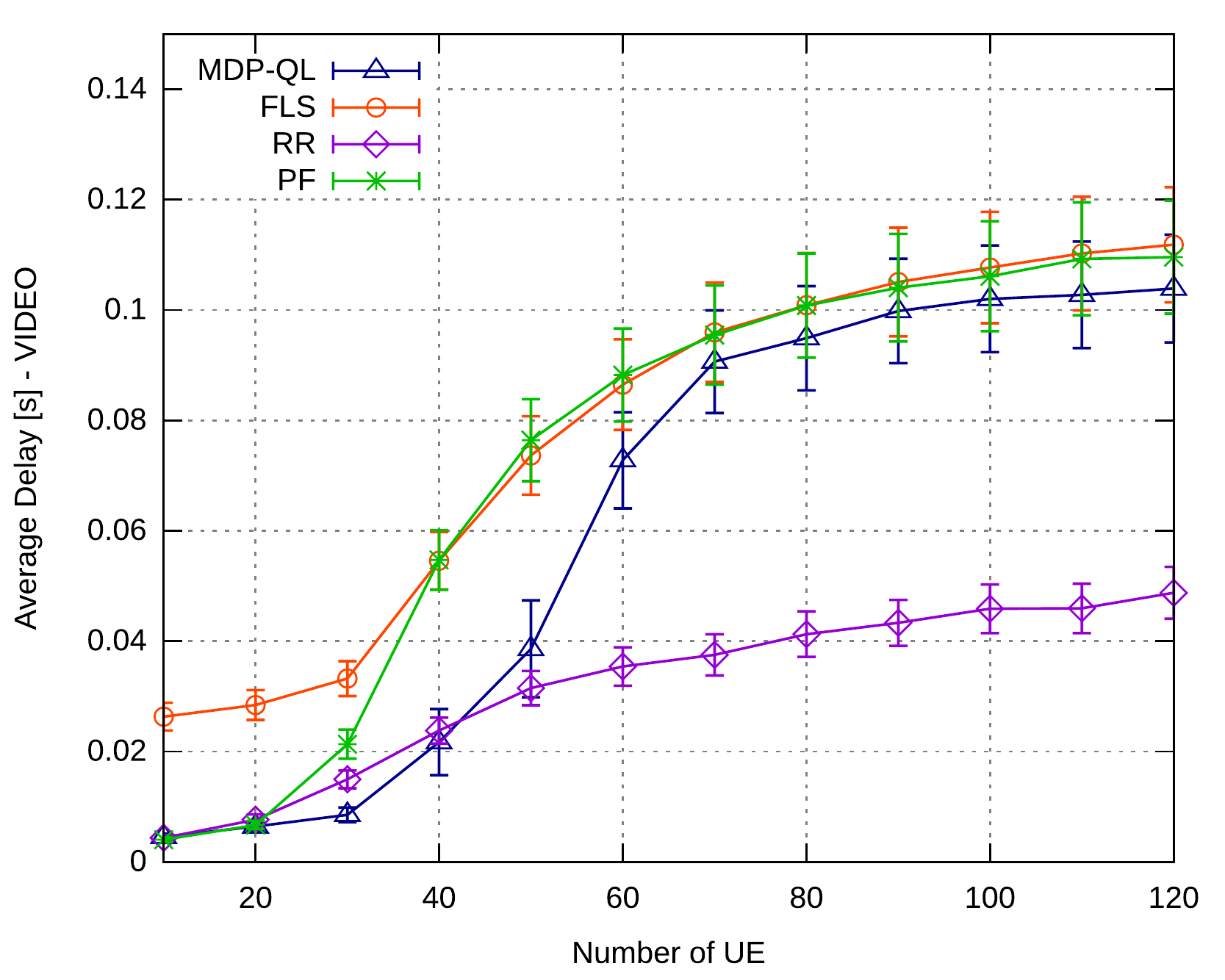}
		\caption{Average delay for Video application}
		\label{delay_video}
	\end{figure}

	Figure \ref{delay_video} presents the values for average delay. The average delay for MDP-QL is not so efficient from 50 UEs in the system. Although, all algorithms maintain delay under the limit for the considered Video application.

\begin{figure}[!htb]
		\centering
		\includegraphics[width=0.4\textwidth]{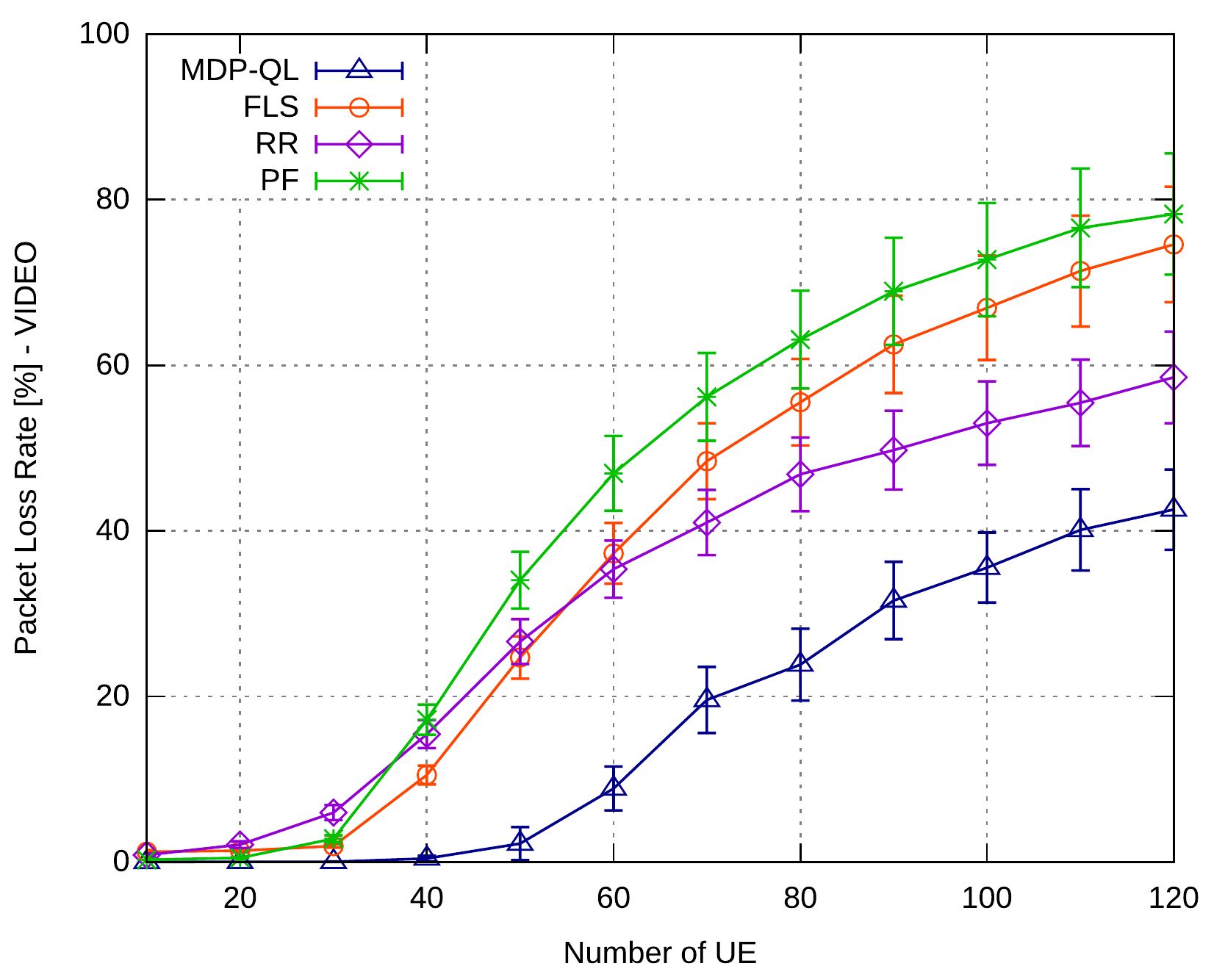}
		\caption{Packet loss rate for Video application}
		\label{plr_video}
	\end{figure}

	A significant result is displayed in Figure \ref{plr_video}. Keeping low packet loss rate is essential to guarantee QoS for real-time traffic. So, Figure \ref{plr_video} shows that MDP-QL can still preserve some QoS level for Video up to 50 UEs. From 50 UEs, the mechanism still maintains the packet loss rate below the achieved by other algorithms.

\begin{figure}[!htb]
		\centering
		\includegraphics[width=0.4\textwidth]{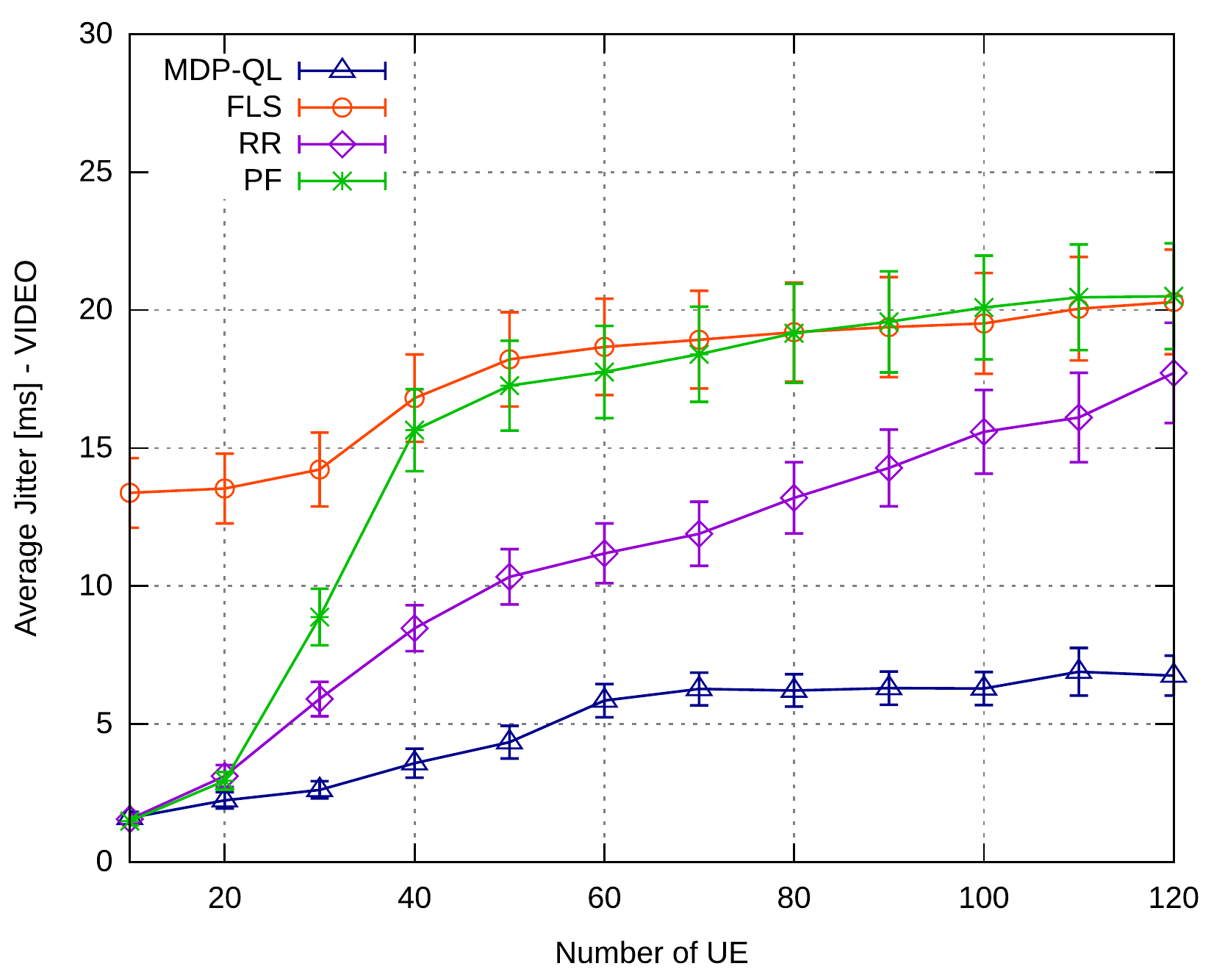}
		\caption{Average jitter for Video application}
		\label{jitter_video}
	\end{figure}

	The average jitter is presented in Figure \ref{jitter_video}. The MDP-QL is quite stable regarding jitter, exhibiting values below all the other algorithms for almost all evaluated scenarios.

\begin{figure}[!htb]
		\centering
		\includegraphics[width=0.4\textwidth]{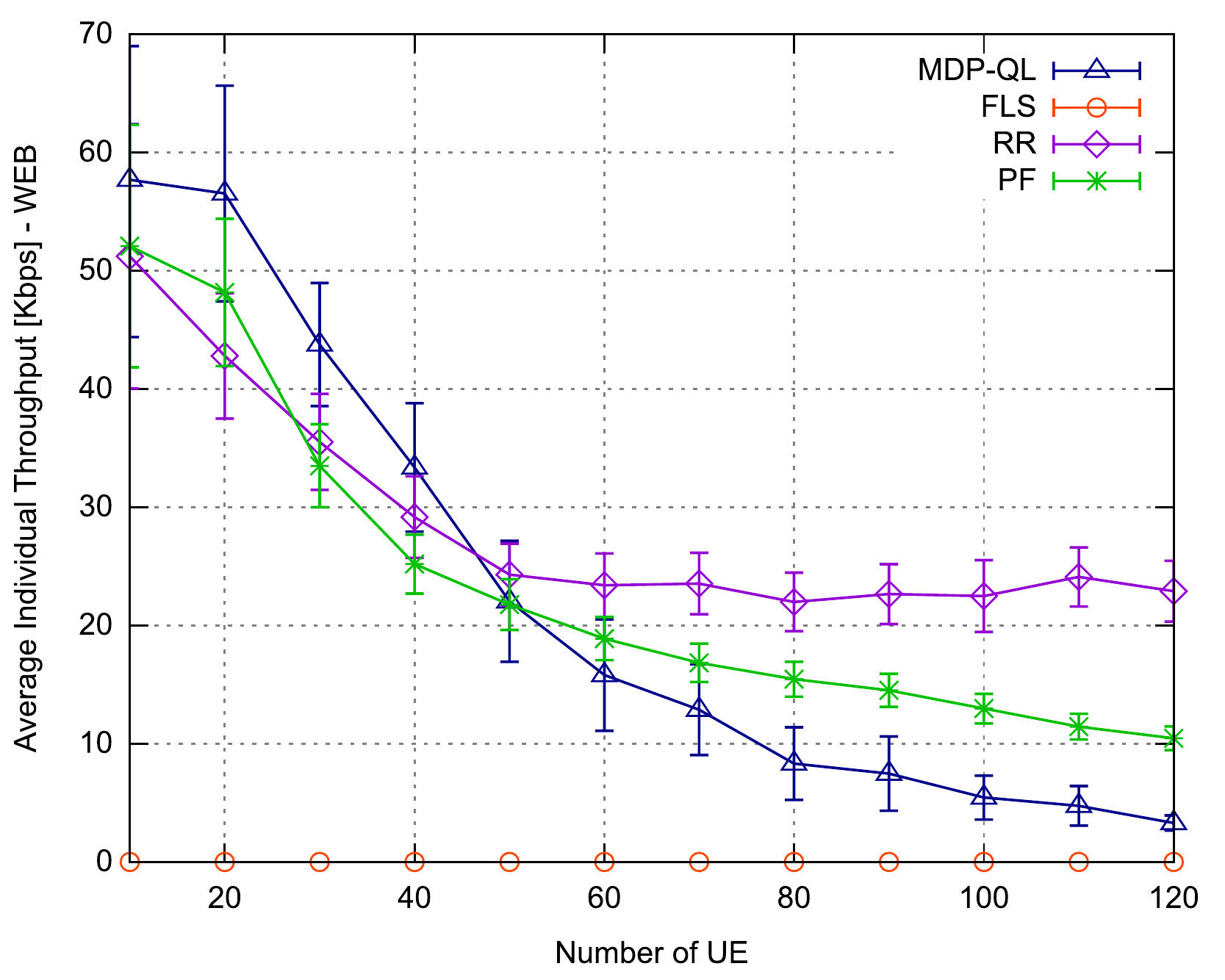}
		\caption{Average throughput for Web application}
		\label{average_throughput_web}
	\end{figure}

	Finally, the Figure \ref{average_throughput_web} presents the results of average throughput obtained for the Web application. It is noticeable the FLS does not meet the application demand even with low traffic load, which indicates that all the effort of the algorithm is directed to support multimedia application in detriment of the best effort traffic. On the other hand, the MDP-QL can support a proper service level, given the adaptive behavior of the reinforcement learning technique.

	Indeed, the proposed mechanism is capable of improving the operation of real-time Video application. It achieves good QoS levels for Video but preserving performance for VoIP and maintaining basic service level for Web application. As the mechanism is based on a technique that aims to optimize some given task, it is expected at least that the proposal can ameliorate some aspects of resource allocation in LTE-A networks.

\section{Conclusion}
\label{conclusion}

	Hitherto was presented a simple reinforcement learning mechanism employed into resource allocation in wireless networks, more specifically the LTE-A technology. The proposed mechanism applies MDP to model the problem of selecting and restricting the number of resources available for each traffic class. In addition, it implements Q-Learning in order to solve and enhance the proposed model.

	Simulation results show good performance measured for Video application that was achieved by the mechanism, which evidences its importance in offering QoS. The mechanism also reaches favorable levels of packet loss rate, average throughput, and average jitter, standing out the proposal in comparison with the analyzed algorithms.

	It would be interesting take advantage of the proposed mechanism integrating it with some technique employed for traffic classification. It is expected therefore some performance improvement with a better traffic classification procedure.

\bibliographystyle{ieeetr}
\bibliography{sbc-template}

\end{document}